\documentclass{article}

\usepackage{arxiv}

\usepackage[utf8]{inputenc} 
\usepackage[T1]{fontenc}    
\usepackage{hyperref}       
\usepackage{url}            
\usepackage{booktabs}       
\usepackage{amsfonts}       
\usepackage{nicefrac}       
\usepackage{microtype}      
\usepackage{lipsum}		
\usepackage{graphicx}
\usepackage{natbib}
\usepackage{doi}

\usepackage{amsmath}
\usepackage{amssymb}
\usepackage{algorithm}
\usepackage{algpseudocode}
\usepackage{tikz}
\usepackage{graphicx,subfigure}
\usepackage{makecell}
\usetikzlibrary{shapes.arrows}
\usetikzlibrary{calc}
\usepackage{array,multirow}
\DeclareMathOperator*{\minimize}{minimize}
\DeclareMathOperator*{\argmin}{argmin}
\usepackage{multicol}
\usetikzlibrary{decorations.pathreplacing,calligraphy}
\usetikzlibrary {arrows.meta}

\title{Distributed Solution of the Inverse Rig Problem in Blendshape Facial Animation}

\date{March 2023}	

\author{ \href{https://orcid.org/0000-0002-5656-9189}{\includegraphics[scale=0.06]{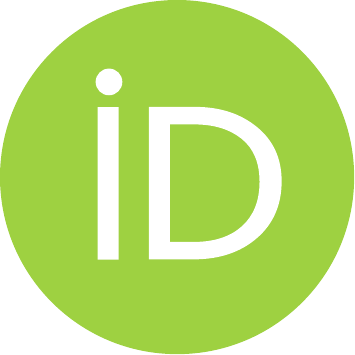}\hspace{1mm}Stevo Racković} \\
	Department of Mathematics\\
	Instituto Superior Técnico\\
	Lisbon, Portugal \\
	\texttt{stevo.rackovic@tecnico.ulisboa.pt} \\
	\And
	\href{https://orcid.org/0000-0003-3071-6627}{\includegraphics[scale=0.06]{orcid.pdf}\hspace{1mm}Cláudia Soares} \\
	Department of Computer Sceince\\
	NOVA School of Science and Technology\\
	Caparica, Portugal \\
	\And
	\href{https://orcid.org/0000-0003-3497-5589}{\includegraphics[scale=0.06]{orcid.pdf}\hspace{1mm}Dušan Jakovetić} \\
	Department of Mathematics\\
	University of Novi Sad\\
	Novi Sad, Serbia \\
}

\renewcommand{\shorttitle}{Distributed Solution of the Inverse Rig}

\hypersetup{
pdftitle={Distributed Solution of the Inverse Rig Problem in Blendshape Facial Animation},
pdfauthor={Stevo Racković, Cláudia Soares, Dušan Jakovetić}
}

\begin{document}
\maketitle

\begin{abstract}
    The problem of rig inversion is central in facial animation as it allows for a realistic and appealing performance of avatars. With the increasing complexity of modern blendshape models, execution times increase beyond practically feasible solutions. A possible approach towards a faster solution is clustering, which exploits the spacial nature of the face, leading to a distributed method. In this paper, we go a step further, involving cluster coupling to get more confident estimates of the overlapping components. Our algorithm applies the Alternating Direction Method of Multipliers, sharing the overlapping weights between the subproblems. The results obtained with this technique show a clear advantage over the naive clustered approach, as measured in different metrics of success and visual inspection. The method applies to an arbitrary clustering of the face. We also introduce a novel method for choosing the number of clusters in a data-free manner. The method tends to find a clustering such that the resulting clustering graph is sparse but without losing essential information. Finally, we give a new variant of a data-free clustering algorithm that produces good scores with respect to the mentioned strategy for choosing the optimal clustering.
\end{abstract}


\section{Introduction}

Blendshape animation is a method in computer graphics, specifically popular for modeling a human face, that animates a 3D mesh $\textbf{b}_0\in\mathbb{R}^{3n}$ by linearly interpolating between a set of predefined morph targets (blendshapes) $\textbf{b}_1,...,\textbf{b}_m\in\mathbb{R}^{3n}$, where $n$ is the number of mesh vertices (Pighin et al., 1998 \cite{Pighin1998SynthesizingRF}, and Lewis et al., 2014 \cite{lewis2014practice}). These morph targets represent different shapes the mesh can take on, and by blending them, a wide range of shapes can be generated. This can be represented as a weighted sum of the morph targets, where the weights $\textbf{w} = [w_1,...,w_m]$ define the amount of influence each morph target has on the final shape
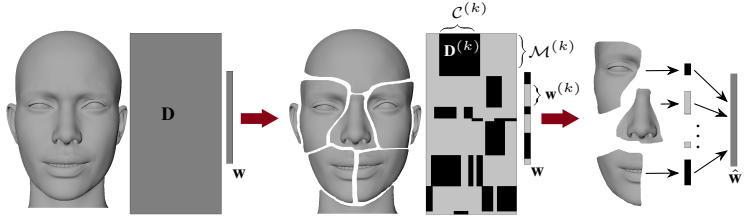
\begin{figure}
    \centering
    \begin{tikzpicture}
    \node[above right, inner sep=0] (image) at (0,0.1){\includegraphics[width=0.6\linewidth]{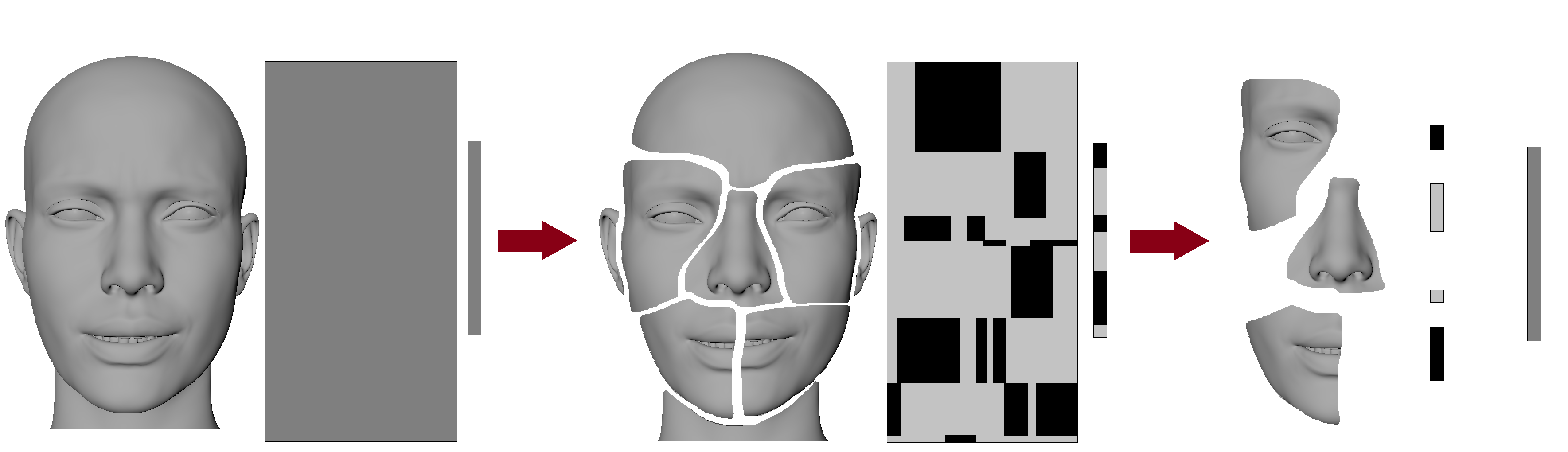}};
    \begin{scope}[
        x={($0.1*(image.south east)$)},
        y={($0.1*(image.north west)$)}]
      \node[black] at (2.2,5) {\scriptsize $\textbf{D}$};
      \node[black] at (3.15,2.25) {\scriptsize $\textbf{w}$};
      \node[white] at (6.15,7.8) {\tiny $\textbf{D}^{(k)}$};
      \draw [decorate, decoration = {calligraphic brace}] (5.85,8.55) --  (6.3,8.55);
      \node[black] at (6.25,9.5) {\tiny $\mathcal{C}^{(k)}$};
      \draw [decorate, decoration = {calligraphic brace}] (6.9,8.35) --  (6.9,6.9);
      \node[black] at (7.35,7.75) {\tiny $\mathcal{M}^{(k)}$};
      \node[black] at (7.15,2.2) {\scriptsize $\textbf{w}$};
      \draw [decorate, decoration = {calligraphic brace}] (7.1,6.1) --  (7.1,5.2);
      \node[black] at (7.5,5.85) {\tiny $\textbf{w}^{(k)}$};
      \node[black] at (9.8,1.95) {\scriptsize $\hat{\textbf{w}}$};
       \draw[-{Stealth}] (8.6,2.1)   -- (9.05,2.1);
       \draw[-{Stealth}] (9.25,2.1)   -- (9.7,3.1);
       \draw[-{Stealth}] (9.25,5.2)   -- (9.7,4.6);
       \draw[-{Stealth}] (8.8,5.2)   -- (9.05,5.2);
       \draw[-{Stealth}] (9.25,6.7)   -- (9.7,5.25);
       \draw[-{Stealth}] (8.6,6.7)   -- (9.05,6.7);
      \node[black] at (9.3,4.1) {\small $\vdots$};

    \end{scope}
    \end{tikzpicture}
    \caption{\textit{Solving the inverse problem in a clustered setting. A blendshape offset matrix $\textbf{D}$ is clustered/segmented in both mesh (rows) and controller (columns) space, so the whole face model is divided into several submodels. The inverse rig problem is solved for each local cluster and the final results are aggregated into the prediction $\hat{\textbf{w}}$. Face model: \textcopyright unrealengine.com/en-US/eula/mhc.}}
    \label{fig:scheme}
\end{figure}
\begin{equation}
    f_L(\textbf{w};\textbf{B}) = \textbf{b}_0 + \sum_{i=1}^m w_i\textbf{b}_i =  \textbf{b}_0 + \textbf{B}\textbf{w}.
\end{equation}
Here, $\textbf{B}\in\mathbb{R}^{3n\times m}$ is a blendshape matrix created by stacking the blendshape vectors as its columns. The weights are then animated over time to produce the desired shape transitions. In modern facial animation, with large $n$ and $m$, a linear model is not sufficient to produce the desired realism, in the first place due to the conflicting deformations --- some pairs of blendshapes $\textbf{b}_i$ and $\textbf{b}_j$ might produce artifacts in the mesh when activated together, hence a corrective blendshape $\textbf{b}^{\{ij\}}\in\mathbb{R}^{3n}$ needs to be sculpted and included with a product of their weights $w_iw_j$ whenever the two are activated simultaneously. Similar holds for combinations of three or more blendshapes, invoking the corrective terms of higher-level, as explained in \cite{rackovic2022CD}. A model with three levels of corrections is defined as 
\begin{equation}
    \begin{split}
        & f_Q(\textbf{w}) =  \textbf{b}_0 + \sum_{i=1}^m w_i\textbf{b}_i + \sum_{\{i,j\}\in\mathcal{P}}w_iw_j\textbf{b}^{\{ij\}} + \\
        &  \sum_{\{i,j,k\}\in\mathcal{T}}w_iw_jw_k\textbf{b}^{\{ijk\}} 
        + \sum_{\{i,j,k,l\}\in\mathcal{Q}}w_iw_jw_kw_l\textbf{b}^{\{ijkl\}},
    \end{split}
\end{equation}
where $\mathcal{P},\mathcal{T}$ and $\mathcal{Q}$ stand for tuples of indices (of sizes 2, 3, and 4, respectively) of the blendshapes that invoke corrective terms. Further in this paper, we will drop the subscript $Q$, and assume that a rig function $f(\cdot)$ always incorporates all the available corrective terms.

A common problem, of primary interest in this paper, is the inversion of the rig. I.e., given a target mesh $\widehat{\textbf{b}}\in\mathbb{R}^{3n}$ (obtained as a 3D scan of an actor or a set of markers), find a configuration of the weight vector $\textbf{w}$ that would closely approximate the target. While the data fidelity term $f(\textbf{w})\approx \widehat{\textbf{b}}$ plays a central role, the solution also needs to satisfy given constraints, specifically $\textbf{0}\leq\textbf{w}\leq\textbf{1}$, where the inequalities here are assumed component-wise. Preferably, the solution should also have as few non-zero weights as possible, as it makes it easier for artists to work with the obtained animation later, and less likely to produce artifacts like anatomically incorrect expressions \cite{seol2011artist}. 

Possible approaches to solving the inverse rig problem can be divided into data-based and model-based. Data-based methods neglect the structure of the underlying rig function and rely on large amounts of animated material that are used to train regression models \cite{holden2015learning, holden2016learning, Kim2021DeepLU}. While this can yield good performance, producing enough training data often poses a problem here, as it demands additional time and effort. On the other side, model-based solutions exploit the structure of the rig functions and rely on optimization techniques rather than data \cite{joshi2006learning, ccetinaslan2016position, rackovic2022majorization}. A state-of-the-art model-based solution is given in \cite{rackovic2022CD}, and it solves the problem 
\begin{equation}\label{eq:holistic_objective}
    \minimize_{\textbf{0}\leq\textbf{w}\leq\textbf{1}} \frac{1}{2} \| f(\textbf{w}) - \widehat{\textbf{b}} \|_2^2 + \alpha \textbf{1}^T\textbf{w},
\end{equation}
using coordinate descent, where $\alpha>0$ is a regularization parameter included to enhance the sparsity of the solution.

As pointed out in \cite{rackovic2021Clustering}, the human face (and blendshape model) have local nature, hence most of the vertices are irrelevant for estimating the weights of the majority of the blendshapes. This calls for a segmented model, where objective (\ref{eq:holistic_objective}) is split into subproblems with only relevant weights estimated over each mesh segment (see Fig. \ref{fig:scheme}). Early works suggested splitting the face manually, by inspection, into the upper and lower regions \cite{choe2001analysis}. Yet this is not convenient for modern models with hundreds of blendshapes in the bases, and more sophisticated and automated methods are needed. While different papers propose segmenting the mesh based on the vertex behavior over animated sequences \cite{joshi2006learning, tena2011interactive}, this makes the clusters susceptible to the quality of training data, and unsuitable for model-based approaches in solving the inverse rig. In \cite{seol2011artist}, the mesh regions are painted manually, and then blendshapes are assigned to the corresponding segments. In \cite{romeo2020data} and \cite{rackovic2021Clustering}, mesh clusters are estimated from a given blendshape matrix, and \cite{rackovic2021Clustering} further automatically assigns blendshapes to the relevant segments. While these clustering approaches help reduce the size of problem (\ref{eq:holistic_objective}), and might even have an effect of an additional regularization of the solution, this brings a question of what to do with the weights that are shared between mesh clusters. In \cite{rackovic2021Clustering}, this is solved by simply averaging the values, yet if the coupling between the clusters is included in the optimization process, this could improve the shared estimates. 


\paragraph*{Contributions}
\begin{enumerate}
    \item We formulate a metric that can be used to evaluate the goodness of the blendshape clusters, based on the overall sparsity and the quality of reconstruction of a given clustering, apriori to the fitting phase. This is useful for choosing the optimal number of clusters $K$ in a data-free manner.
    \item We propose an adjustment to the blendshape assignment within the clustering technique of \cite{rackovic2021Clustering}, which, in general, results in a denser graph but a higher reconstruction quality.
    \item We propose a model-based solution to the inverse rig problem in a clustered setting. The proposed method applies the alternating direction method of multipliers (ADMM) \cite{boyd2011ADMM}, in combination with coordinate descent similar to \cite{rackovic2022CD}, allowing coordination between the clusters and adjusted estimates of the shared weights.
\end{enumerate}

This paper follows the pipeline consisting of the above contributions in the following way. Several instantiations of the clusterings are performed and evaluated based on the proposed metric for estimating the trade-off between the error reconstruction and density of the produced segmented blendshape matrix in order to choose the best representative clustering. It is important here to note that, while we propose a new clustering method, this pipeline can work with an arbitrary clustering method, as shown in the results section of the paper. Finally, the clusters are used to solve the inverse rig in a distributed manner, where introducing ADMM allows the coupling of the overlapping components, as opposed to a naive clustered solution that observes each cluster independently. The results show that the pipeline produces solutions closely matching that of the holistic approach in terms of sparsity and accuracy, while significantly reducing the execution time ($50\%$ reduction). A naive clustered solution demands slightly less time than a proposed method, but it does not compare with our solution in either accuracy or the sparsity metric, and supplementary video materials show a clear superiority of our results. The codes will be made available upon the paper acceptance.

\section{Clustering of the face}

The clustering methods of Seol et al., 2011 \cite{seol2011artist} (here termed \textit{SSKLN}, from the initials of the authors) and of Racković et al., 2021 \cite{rackovic2021Clustering} (here termed \textit{RSJD}) transform the blendshape matrix $\textbf{B}\in\mathbb{R}^{3n\times m}$ into a matrix of offset values $\textbf{D}\in\mathbb{R}^{n\times m}$. Columns $\textbf{d}_i$ of this matrix are obtained as offsets for each controller $i$: 
\begin{equation}
{d}_i^l = \big\|[{b}_i^{3l},{b}_i^{3l-1},{b}_i^{3l-2}]\big\|_2^2 \,,\,\text{   for } l=1,...,n.    
\end{equation}
Here $b_i^{3l-2}$ represents entry of a blendshape $\textbf{b}_i$ that corresponds to $x$ coordinate of the vertex $\textbf{v}_l$. Similarly, superscripts $3l-1$ and $3l$ correspond to $y$ and $z$ coordinates of $\textbf{v}_l$. The method proposed in Romeo et al., 2020 \cite{romeo2020data} (here termed \textit{RS}, from the initials of the authors), rearranges blendshape matrix $\textbf{B}\in\mathbb{R}^{3n\times m}$ into a matrix $\Delta\in\mathbb{R}^{n\times 3m}$, whose elements are $\Delta_{i,3l}=b_i^{3l}$ for $l=1,...,n$. I.e., each blendshape $\textbf{b}_i$ is decomposed into three vectors, containing $x, y$, and $z$ coordinates, respectively, and then these vectors are stacked as columns of matrix $\Delta$.

\textit{RSJD} and \textit{RS} perform K-Means clustering \cite{hartigan1979algorithm} over the rows of $\textbf{D}$ and $\Delta$, respectively, to obtain mesh clusters $\mathcal{M}^{(k)}$, for $k=1,...,K$. The \textit{SSKLN} assumes that an artist manually selects the four mesh segments. Further, \textit{SSKLN} assigns each blendshape to a relevant mesh segment using the following procedure. For each blendshape $i$, a magnitude of deformation over the segment $\mathcal{M}^{(k)}$, denoted $s_i^{(k)}$, is computed as the sum of the elements of $\textbf{d}_i$ within the segment. The overall deformation of the blendshape $i$ is a sum of the entire vector, $s_i = \textbf{1}^T\textbf{d}_i = \sum_{k=1}^4s_i^{(k)}$. The controller $i$ is assigned to each mesh cluster $k$ where $s_i^{(k)} > \frac{s_i}{2}$, producing in this way $K=4$ controller clusters $\mathcal{C}^{(k)}$, as illustrated in Fig. \ref{fig:scheme}. In \textit{RSJD}, each column vector $\textbf{d}_i$, of the matrix $\textbf{D}$, is compressed into $\textbf{h}_i\in\mathbb{R}^K$ such that ${h}_i^k = \frac{\sum_{l \in \mathcal{M}^{(k)}}{d}_i^l}{|\mathcal{M}^{(k)}|}\,\,\text{ for }\,k=1,..,K.$ Then K-means is performed over $\textbf{h}_i$ to split it into two subvectors, one with high entries and the other one with low. The controller $i$ is assigned to all the mesh clusters corresponding to the high-valued labels. The number of clusters $K$ is left as a user-defined parameter and is chosen based on the performance over training data. 

In this Section, we propose a data-free method for selecting a good value of $K$ --- we want the resulting clustering to produce a relatively sparse reconstruction of the blendshape matrix without losing important information.

Recall that, while the mesh segmentation of the \textit{RSJD} and \textit{RS} is in a way similar, \textit{RS} does not provide a method for blendshape assignment to the mesh clusters. Hence we augment it in this paper, for the purpose of solving the inverse rig, by applying the same method as in the \textit{RSJD}.


\paragraph*{Proposed Clustering Method.}

\begin{figure}
    \centering
    \includegraphics[width=0.6\linewidth]{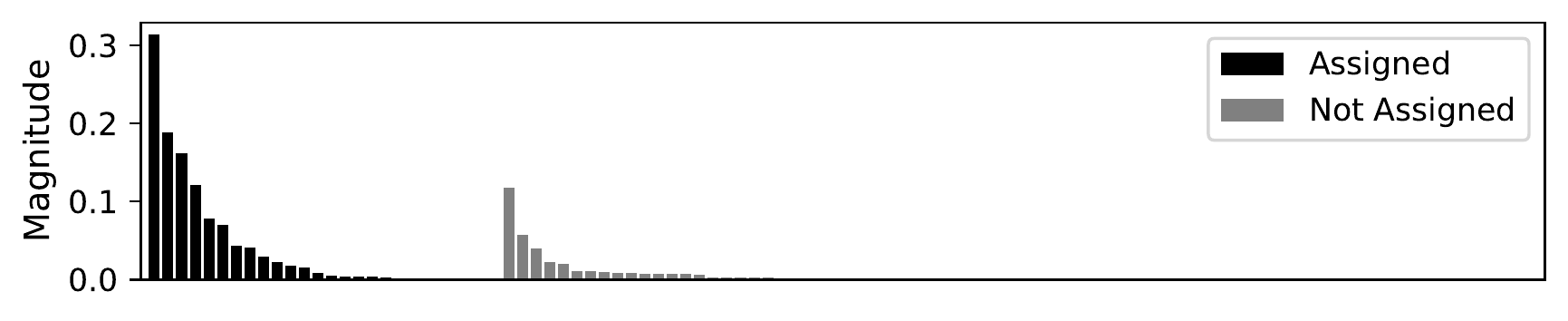}
    \caption{\textit{Average magnitude of deformation produced by each blendshape in a chosen cluster, obtained by the method of \textit{RSJD}.}}
    \label{fig:magnitudes}
\end{figure}

The \textit{RSJD} method clusters the face model in both mesh and blendshape space, in a data-free manner, relying purely on a blendshape matrix. While the presented results show great performance, there is an issue in  the blendshape assignment that we want to address. Blendshapes are assigned to the mesh segments where their effect is significantly larger than in the others, yet it does not imply that their effect within the corresponding cluster will be significant compared to other blendshapes. In particular, it might happen that, within a specific mesh segment, there are blendshapes non-assigned to it but whose overall magnitude of deformation is significantly larger than some of the assigned ones (Fig. \ref{fig:magnitudes}). 

In this paper, we propose a simple adjustment --- the lowest magnitude value among all the blendshapes initially assigned to an observed cluster is taken as a threshold, $p^{(k)}=\min \sum_{i\in\mathcal{M}^{(k)}}(b_j^i)^2$ for $j\in\mathcal{C}^{(k)}$. Consequently, all the other blendshapes whose deformation magnitude is larger than the threshold, i.e., such that $\sum_{i\in\mathcal{M}^{(k)}}(b_l^i)^2>p^{(k)}$ for $l\not\in\mathcal{C}^{(k)}$, are assigned to the cluster as well. This method will be termed \textit{RSJD}\textsubscript{A} ("\textit{A}" standing for "\textit{adjusted}") throughout the paper.

\paragraph*{Choosing the Number of Clusters $K$.}\label{sec:choosingK}

\begin{figure}
\centering
    \begin{tikzpicture}
    \node[above right, inner sep=0] (image) at (0.5,0){
    \includegraphics[width=0.6\linewidth]{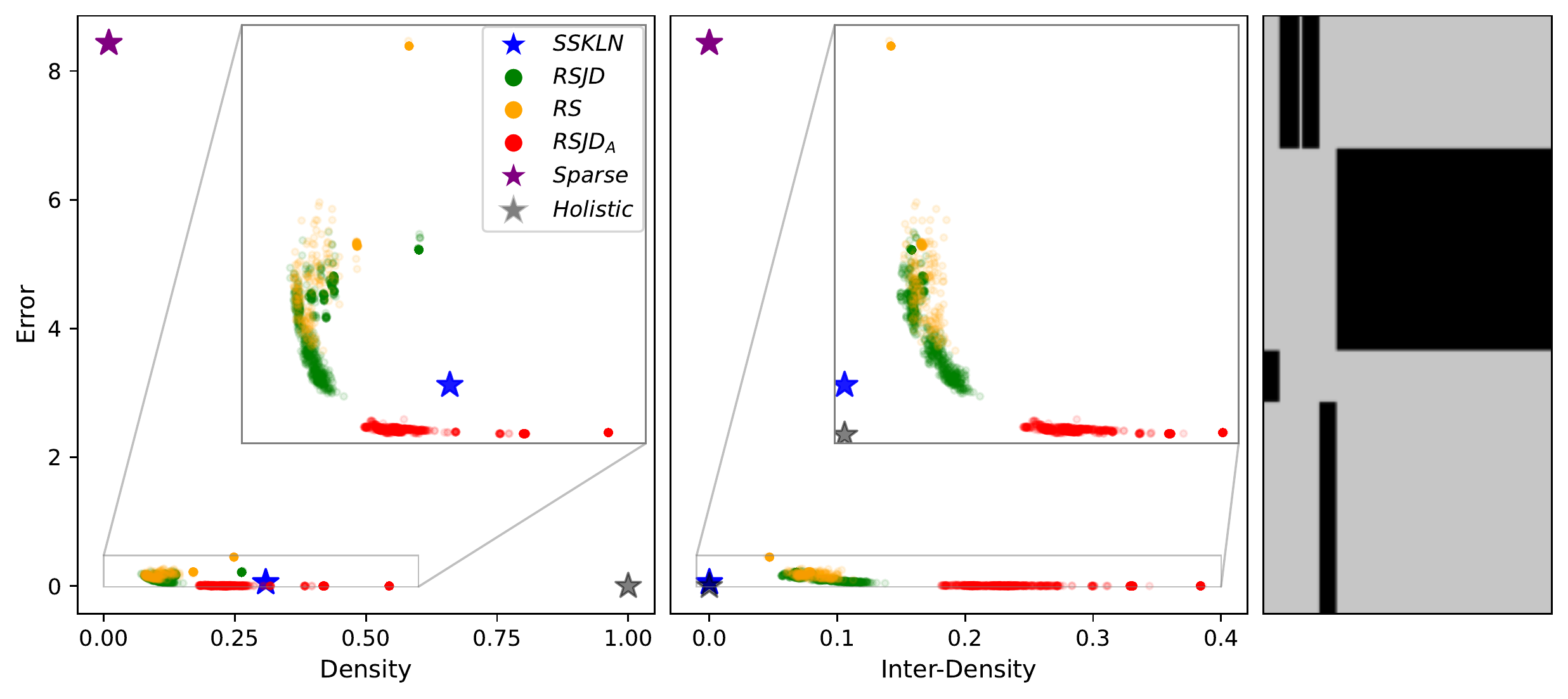}};
    \begin{scope}[
        x={($0.1*(image.south east)$)},
        y={($0.1*(image.north west)$)}]
      \node[white] at (9,6) {\tiny{ $\textbf{B}\textsuperscript{(k)}$} };
      \node[black] at (8.05,6) {\tiny $\bar{\textbf{B}}\textsuperscript{(k)}$ };
    \end{scope}
    \end{tikzpicture}
    \caption{\textit{Left: Trade-off between the density ($E_D$) and the reconstruction error ($E_R$) of the clustered blendshape matrix, for different clusterings. Middle: Trade-off between the inter-density ($E_{ID}$) and the reconstruction error ($E_R$) of the clustered blendshape matrix, for different clusterings. Each dot represents a single clustering output, with $K$ taking values from $4$ to $m=102$ for the \textit{RSJD}, \textit{RS} and \textit{RSJD}\textsubscript{A} (and fixed to $K=4$ for the \textit{SSKLN}, and $K=m=102$ for \textit{Sparse}). Right: Blendshape matrix clustered using the \textit{SSKLN}. Dark entries correspond to clusters, and light entries are discarded vertex-blendshape pairs.}}
    \label{fig:ClustersTradeoff}
\end{figure}

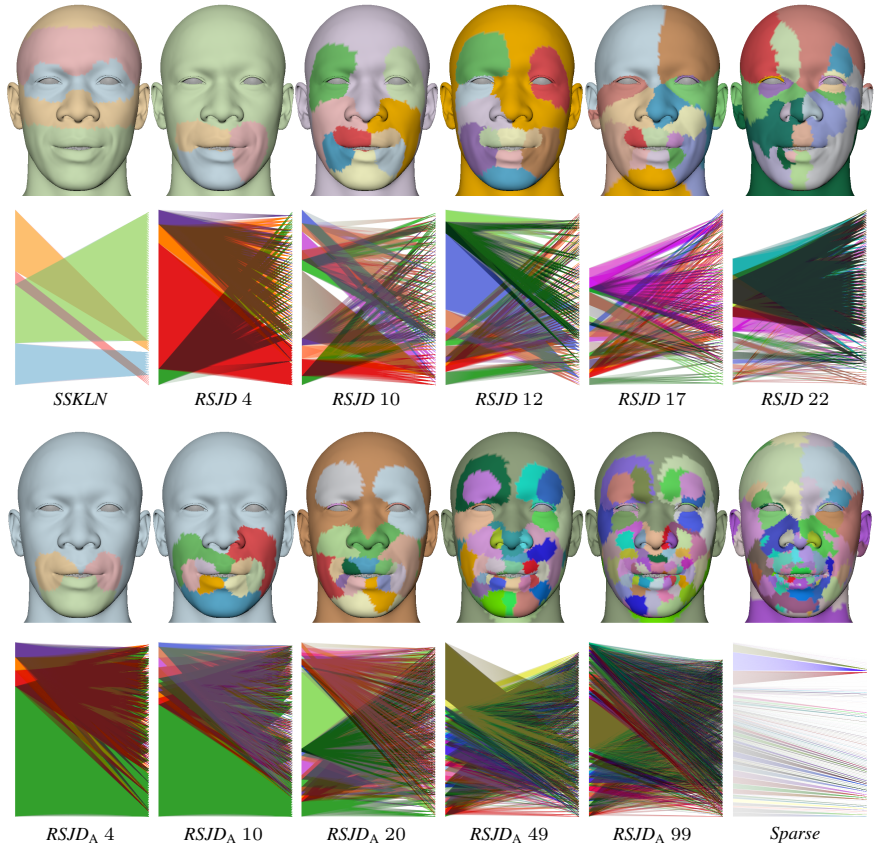
\begin{figure}
    \centering
    \begin{tikzpicture}
    \node[above right, inner sep=0] (image) at (2,8.5){
        \includegraphics[width=0.7\linewidth]{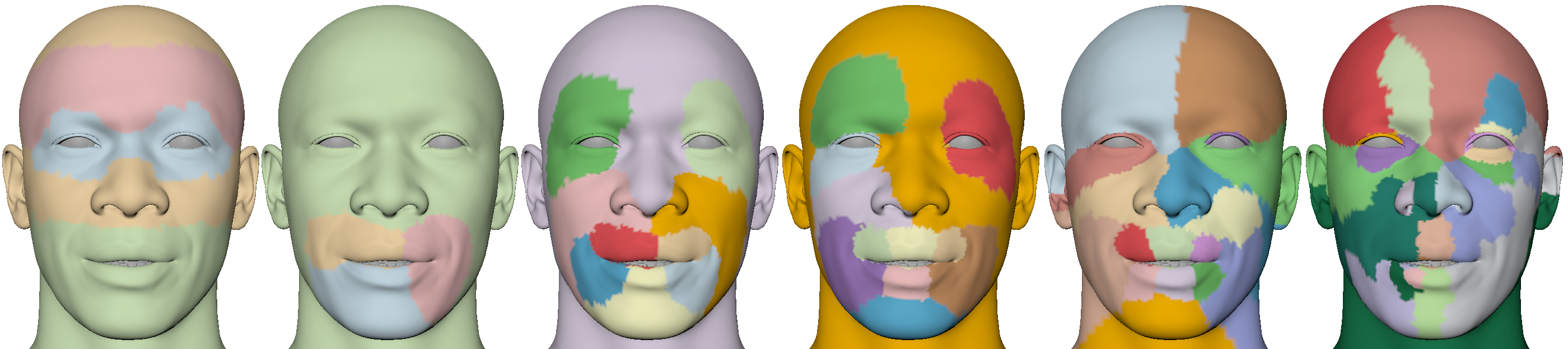}};
    \node[above right, inner sep=0] (image) at (2,5.9){
        \includegraphics[width=0.7\linewidth]{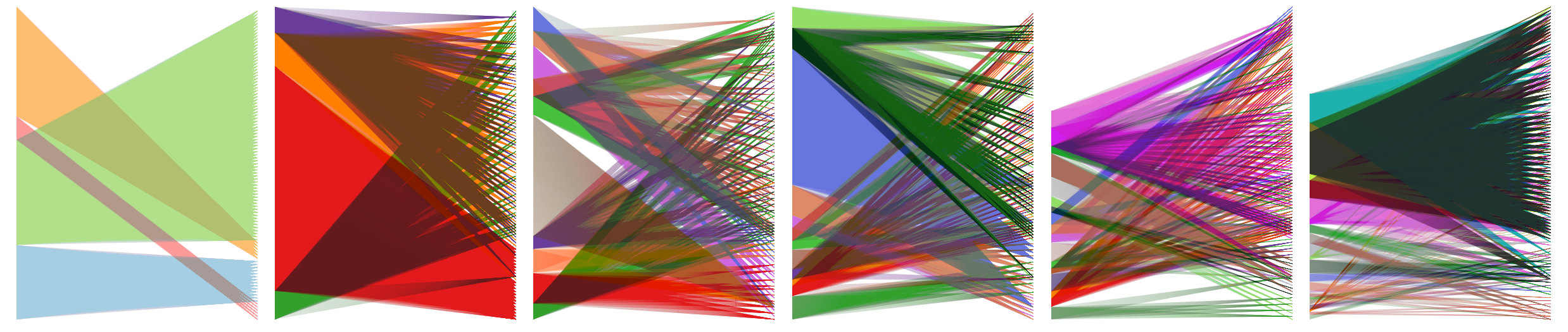}};
    \node[above right, inner sep=0] (image) at (2,2.8){
        \includegraphics[width=0.7\linewidth]{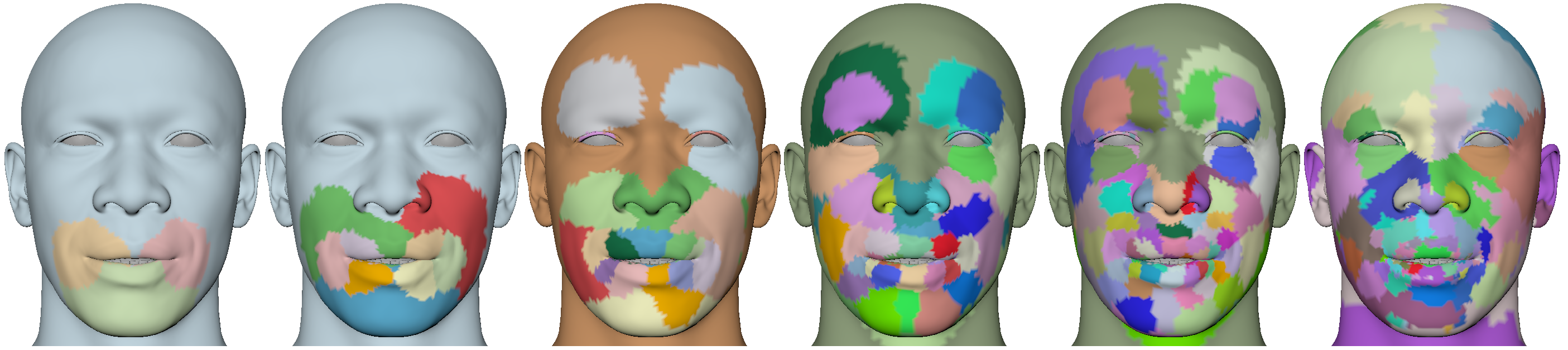}};
    \node[above right, inner sep=0] (image) at (2,0.2){
        \includegraphics[width=0.7\linewidth]{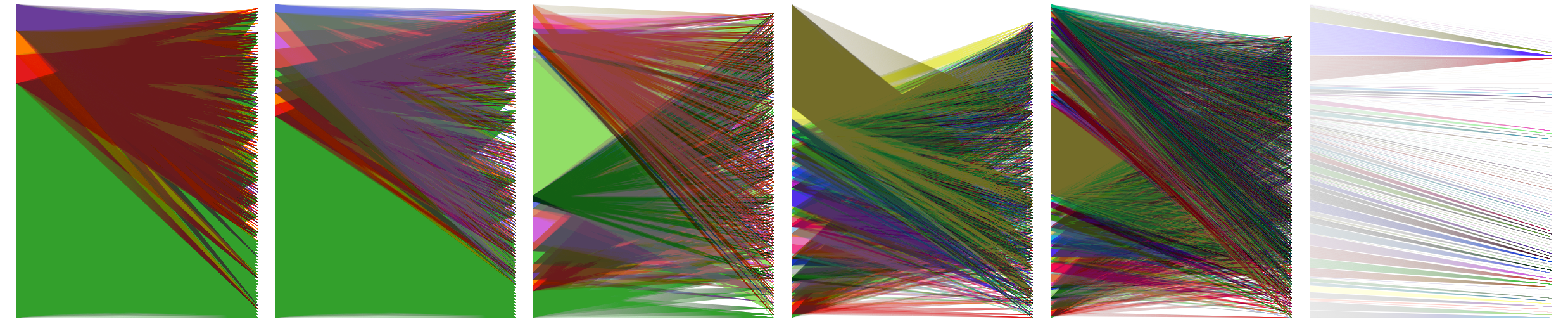}};
    \node[black] at (3.0,0) {\scriptsize \textit{RSJD}\textsubscript{A} $4$};
    \node[black] at (4.9,0) {\scriptsize \textit{RSJD}\textsubscript{A} $10$};
    \node[black] at (6.8,0) {\scriptsize \textit{RSJD}\textsubscript{A} $20$};
    \node[black] at (8.7,0) {\scriptsize \textit{RSJD}\textsubscript{A} $49$};
    \node[black] at (10.6,0) {\scriptsize \textit{RSJD}\textsubscript{A} $99$};
    \node[black] at (12.5,0) {\scriptsize \textit{Sparse}};
    \node[black] at (3.0, 5.8) {\scriptsize \textit{SSKLN}};
    \node[black] at (4.9, 5.8) {\scriptsize \textit{RSJD} $4$};
    \node[black] at (6.8, 5.8) {\scriptsize \textit{RSJD} $10$};
    \node[black] at (8.7, 5.8) {\scriptsize \textit{RSJD} $12$};
    \node[black] at (10.6, 5.8) {\scriptsize \textit{RSJD} $17$};
    \node[black] at (12.5, 5.8) {\scriptsize \textit{RSJD} $22$};    
    \end{tikzpicture}
    \caption{\textit{Clustering outputs of the four approaches --- for the \textit{RSJD} and \textit{RSJD}\textsubscript{A}, the number of clusters $K$ is indicated. Besides the mesh clusters, the figure shows bipartite graphs consisting of the vertices (left-hand-side) and controllers (right-hand-side). Each color indicates a single cluster, with edges representing the cluster correspondences. The avatar \textit{Jesse} is acquired from the MetaHuman Creator (\textcopyright unrealengine.com/en-US/eula/mhc).}}
    \label{fig:MeshCLusters}
\end{figure}

  Let us consider a blendshape matrix $\textbf{B}\in\mathbb{R}^{3n\times m}$, segmented into submatrices $\textbf{B}^{(k)}\in\mathbb{R}^{3n_k\times m_k}$, for $k=1,...,K$, as illustrated in Fig. \ref{fig:ClustersTradeoff} (right). The \textit{Density} of the clustering represents the percentage of the elements of the blendshape matrix kept after the clustering, and it can be computed as $E_D = \sum_{k=1}^K\frac{n_{k} m_{k}}{nm}$, where $n_k=|\mathcal{M}^{(k)}|<m$ and $m_k=|\mathcal{C}^{(k)}|<m$ are the number of vertices and the number of blendshapes assigned to cluster $k$, respectively. We can also understand this as a number of edges $E$ in a bipartite graph $ G=(U,V,E) $ where $U$ represents all the vertices of the mesh and $V$ the controllers --- an edge $(i,j)\in E$ is drawn for every $i\in\mathcal{M}^{(k)}$ and $j\in\mathcal{C}^{(k)}$, for $k=1,...,K$ (see Fig. \ref{fig:MeshCLusters}). 
  
  While $E_D$ shows the overall density of the model, we are also interested in the size of the clusters' overlap. We call this \textit{Inter-Density}, $E_{ID}$. It represents the number of edges shared between multiple clusters in the bipartite graph $G$, that is, edges $(i,j)\in E$ such that $i\in\mathcal{M}^{(k_1)}\cup \mathcal{M}^{(k_2)}$ and $j\in\mathcal{C}^{(k_1)}\cap \mathcal{C}^{(k_2)}$ for some $k_1, k_2 = 1,...,K$, and $k_1\neq k_2$. This metric will indicate how much coupling should be added between the clusters in the fitting phase. 
  
  As a heuristic for measuring the \textit{Reconstruction Error}, we will focus on the ratio between the kept and dismissed elements of the blendshape matrix. Let us observe the submatrices $\bar{\textbf{B}}^{(k)}\in\mathbb{R}^{3n_k\times (m-m_k)}$, for $k=1,...,K$, which represent rejected elements of $\textbf{B}$. Compute the sum of the squared entries of all these matrices $E_{R1} = \sum_{k=1}^K\sum_{i=1}^{n_k}\sum_{j=1}^{m-m_k} (\bar{B}^{(k)}_{ij})^2$, and a sum of the kept elements as $E_{R2} = \sum_{k=1}^K\sum_{i=1}^{n_k}\sum_{j=1}^{m} (B^{(k)}_{ij})^2$. The reconstruction error is now computed as $E_R=E_{R1}/E_{R2}$. 

  The two metrics, $E_D$ and $E_R$, are, in general, inversely proportional, and an optimal clustering would exhibit relatively small values of each. This trade-off is illustrated in Fig. \ref{fig:ClustersTradeoff} (left). The holistic case (i.e., the blendshape matrix without clustering) will always have $E_R=0$ and $E_D=1$, as illustrated with a gray star. Presented are also the results of the four clustering methods \textit{SSKLN}, \textit{RSJD}, and \textit{RS} and \textit{RSJD}\textsubscript{A}. The \textit{SSKLN} is represented by a single scatter (blue star), as it is deterministic in the sense that the four clusters are selected as suggested in \cite{seol2011artist}. The other three can vary in terms of the number of clusters $K$ and for different repetitions of the same $K$; hence, we repeat the clustering 1000 times, with $K$ taking values between $4$ and $m=102$. For the sake of completeness, we also consider the extremely sparse case --- where each mesh vertex is assigned to exactly one blendshape controller, choosing always the one producing the largest offset. This is termed \textit{Sparse}, and presented by the purple star in Fig. \ref{fig:ClustersTradeoff}. The instances of the \textit{RSJD} are closer to the lower left corner (of the left subfigure) than \textit{SSKLN} or \textit{RSJD}\textsubscript{A}, however, we need to zoom in to get a better idea of the relationships. The \textit{RSJD}\textsubscript{A}  in general leads to quite low $E_R$, but $E_D$ can get relatively large, while the \textit{RSJD} is of lower density but higher $E_R$. The \textit{SSKLN} is suboptimal in this plot. However, a relationship between $E_R$ and $E_{ID}$ does not need to follow the same shape (Fig. \ref{fig:ClustersTradeoff}, middle). In this case, the distinction between the \textit{RSJD} and \textit{RSJD}\textsubscript{A} is even cleaner, however, notice that the \textit{SSKLN} has $E_{ID}=0$ as its clusters have no overlapping. In both plots, the \textit{RS} closely follows the behavior of the \textit{RSJD}, hence we will eliminate it from further consideration.

An optimal choice of the clustering (and hence $K$) should be based on these plots, choosing the point near the elbow of the trade-off curve, for each of the approaches. For \textit{RSJD}\textsubscript{A} approach, this would be one of the clusterings with the lowest $E_D$, and for the \textit{RSJD}, the one with low $E_R$. We proceed to work on several different choices of $K$ for the two approaches. We will show later in the results sections that a standard procedure of cross-validation, as used in prior works, leads to the same conclusions on the choice of $K$, validating that the considered $K$ selection works. In Fig. \ref{fig:MeshCLusters} (accompanied by Table \ref{tab:clusters_table}) we show mesh clusters and bipartite graphs between the mesh vertices (left-hand-side) and the blendshapes (right-hand-side), colored corresponding to the cluster assignment. One can see that, in general, the lower number of clusters leads to a more dense graph.

\begin{table}
    \centering
    \small
    \caption{\textit{Values of the clustering for each of the four methods.}}\label{tab:clusters_table}
    \begin{tabular}{|c|c|c|c|c|c|c|c|c|c|c|c|c|c|}
                   \hline
               & \textit{SSKLN} & \multicolumn{5}{c|}{\textit{RSJD}} & \multicolumn{5}{c|}{\textit{RSJD}\textsubscript{A}}  & \textit{Sparse} \\
                   \hline
   $K$   & $4$    & $4$     & $10$    & $12$   & $17$   & $22$   & $4$    & $10$    & $20$    & $50$   & $102$  & $102$   \\
   $E_R$ & $0.057$& $0.214$ & $0.159$ & $0.138$& $0.089$& $0.035$& $0.002$& $0.001$ & $0.004$ & $0.003$& $0.005$& $8.434$ \\
   $E_D$ & $0.308$& $0.262$ & $0.098$ & $0.110$& $0.086$& $0.148$& $0.544$& $0.421$ & $0.253$ & $0.237$& $0.208$& $0.009$ \\
$E_{ID}$ & $0.0$  & $0.068$ & $0.056$ & $0.071$& $0.072$& $0.136$& $0.383$& $0.330$ & $0.252$ & $0.237$& $0.208$& $0.0$   \\
                   \hline
    \end{tabular}
\end{table}


\section{Distributed Solution to the Rig Inversion Problem}\label{sec:admm}

\begin{figure*}
    \centering
    \includegraphics[width=\linewidth]{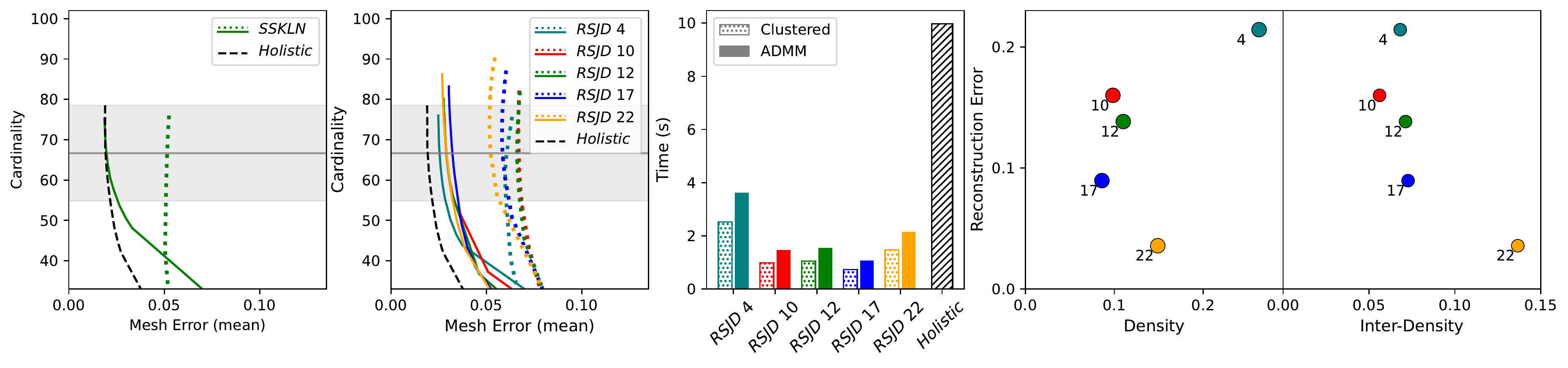}
    \includegraphics[width=\linewidth]{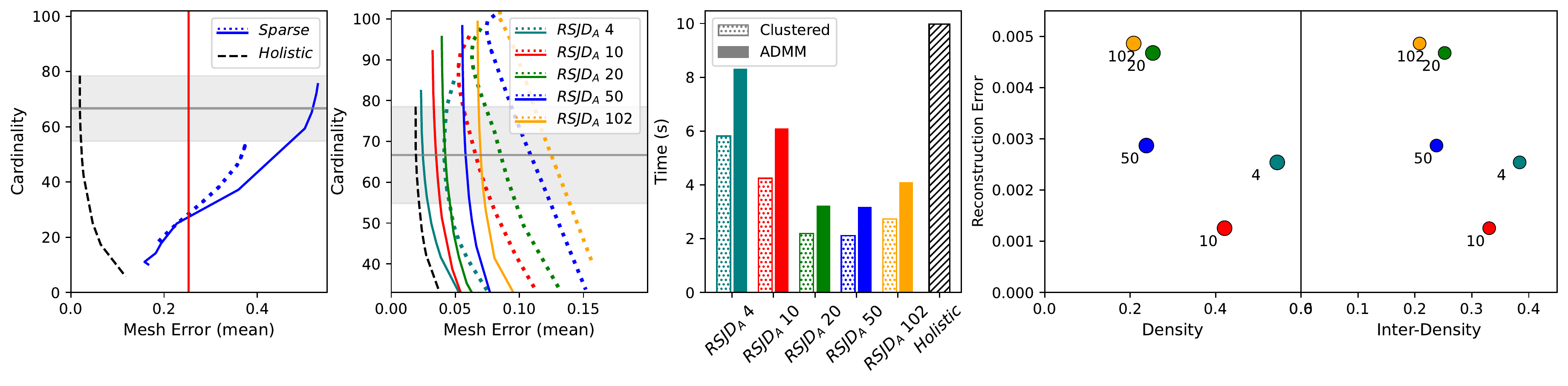}
    \caption{\textit{The first two columns show trade-off curves between the mesh error (RMSE) and cardinality (number of non-zero weights), for different methods, (and different choices of $K$ for the \textit{RSJD} and \textit{RSJD}\textsubscript{A}) as functions of the regularization parameter $\alpha>0$. The dotted lines represent a naive clustered solution, while the solid lines of the same color are the corresponding ADMM solution. The black dashed line shows a holistic approach. The gray horizontal line shows the cardinality of the ground-truth data, with a shaded region marking its standard deviation. A red vertical line in the plot of the \textit{Sparse} approach represents a baseline where at each frame the weight vector is set to $\textbf{w}=\textbf{0}$. For the \textit{RSJD} and \textit{RSJD}\textsubscript{A}, we also represent the average execution time for each choice of $K$ (the third column), as well as the trade-off between $E_R$ and $E_D$ (the fourth column) and $E_R$ and $E_{ID}$ (the last column).}}
    \label{fig:fig_TrainRes}
\end{figure*}

The objective function for the inverse rig problem, as formulated in a state-of-the-art \cite{rackovic2022CD}, is (\ref{eq:holistic_objective}). In the clustered setting, one can simply split this problem into subproblems 
\begin{equation}\label{eq:clustered_objective}
    \minimize_{\textbf{0}\leq\textbf{w}^{(k)}\leq\textbf{1}} \frac{1}{2} \| f^{(k)}(\textbf{w}^{(k)}) - \widehat{\textbf{b}}^{(k)} \|_2^2 + \alpha^{(k)} \textbf{1}^T\textbf{w}^{(k)},
\end{equation}
for $k=1,...,K$, where $\textbf{w}^{(k)}\in\mathbb{R}^{m_k}$ is a vector containing only the $m_k$ weights assigned to the cluster $k$; $\widehat{\textbf{b}}^{(k)}\in\mathbb{R}^{3n_k}$ is a subvector of the target mesh $\widehat{\textbf{b}}$, consisting of the $n_k$ vertices from the corresponding cluster, and similarly, $f^{(k)}(\cdot)$ is a blendshape function restricted only to the vertices and controllers within the cluster $k$  (See Fig. \ref{fig:scheme}).

If these subproblems are solved independently, they yield a set of local weight vectors $\hat{\textbf{w}}^{(k)}$, that should be merged into a single global prediction vector $\hat{\textbf{w}}$. For the controllers that are shared among multiple clusters, the final value is taken as the average of all the estimates. More formally, we introduce the mapping from local variable indices into a global variable index as $j=\mathcal{G}(k,i)$, which means that for some local variable $\textbf{v}^{(k)}$ and a global variable $\textbf{v}$, a local variable component $(\textbf{v}^{(k)})_i$ corresponds to the global variable component $\textbf{v}_j$. We also introduce a diagonal matrix $\textbf{S}\in\mathbb{R}^{m\times m}$ that has entries corresponding to the multiplicity of each controller over the clusters, i.e., $S_{ii} = \sum_{k=1}^K\sum_{\mathcal{G}(k,i)}1$. Now, the global weight estimate is obtained as $\hat{\textbf{w}} = \textbf{S}^{-1} \sum_{k=1}^K\textbf{v}^{(k)}$, where the entries of $\textbf{v}^{(k)}\in\mathbb{R}^m$ are the values of $\hat{\textbf{w}}^{(k)}$ obtained for the corresponding cluster, i.e., $(\textbf{v}^{(k)})_{\mathcal{G}(k,i)} = (\hat{\textbf{w}}^{(k)})_i$.

\paragraph*{Solution via ADMM.}\label{sec:proposedADMM}

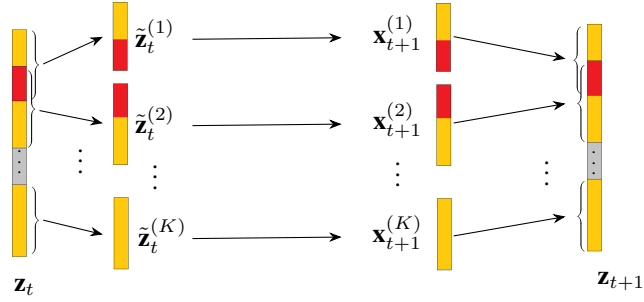
\begin{figure}
    \centering
    \begin{tikzpicture}
    \node[above right, inner sep=0] (image) at (0,0.1){\includegraphics[width=0.6\linewidth]{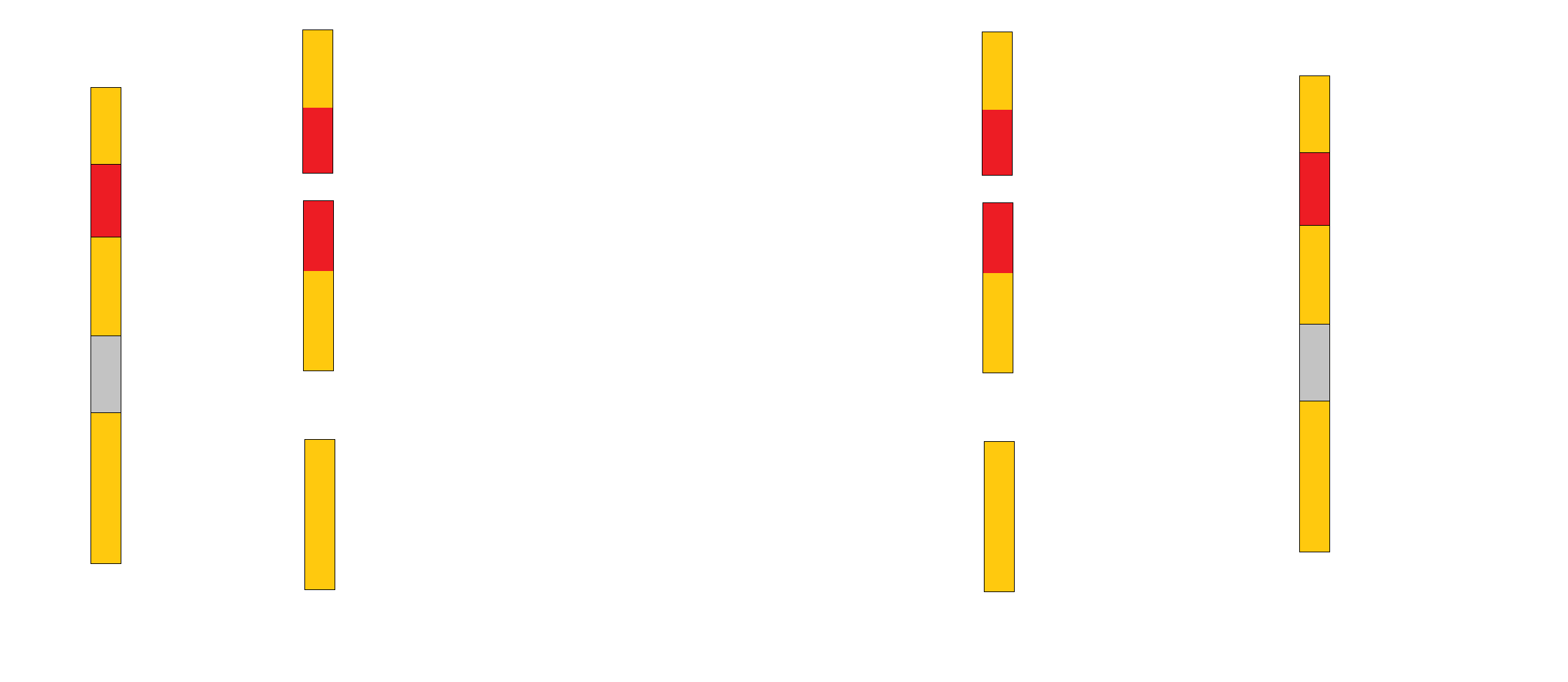}};
    \begin{scope}[
        x={($0.1*(image.south east)$)},
        y={($0.1*(image.north west)$)}]
      \node[black] at (0.75,1) { $\textbf{z}_t$};
      \node[black] at (0.68,5) {\scriptsize $\vdots$};
      \draw [decorate, decoration = {calligraphic brace}] (0.85,8.7) --  (0.85,6.65);
      \draw [decorate, decoration = {calligraphic brace}] (0.8,7.5) --  (0.8,5.2);
      \draw [decorate, decoration = {calligraphic brace}] (0.85,4) --  (0.85,2);
       \draw[-{Stealth}] (1,3.)   -- (1.8,2.55);
       \draw[-{Stealth}] (0.95,6.3)   -- (1.8,5.9);
       \draw[-{Stealth}] (1,7.7)   -- (1.8,8.5);
      \node[black] at (1.5,5) { $\vdots$};
      \node[black] at (2.5,8.5) { $\tilde{\textbf{z}}_t^{(1)}$};
      \node[black] at (2.5,5.9) { $\tilde{\textbf{z}}_t^{(2)}$};
      \node[black] at (2.5,4.5) { $\vdots$};
      \node[black] at (2.6,2.5) { $\tilde{\textbf{z}}_t^{(K)}$};
      \node[black] at (5.75,8.5) { $\textbf{x}_{t+1}^{(1)}$};
      \node[black] at (5.75,5.9) { $\textbf{x}_{t+1}^{(2)}$};
      \node[black] at (5.75,4.5) { $\vdots$};
      \node[black] at (5.75,2.5) { $\textbf{x}_{t+1}^{(K)}$};
       \draw[-{Stealth}] (3,8.4)   -- (5,8.4);
       \draw[-{Stealth}] (3,5.8)   -- (5,5.8);
       \draw[-{Stealth}] (3,2.4)   -- (5,2.4);
       \draw[-{Stealth}] (6.5,8.4)   -- (8,7.7);
      \draw [decorate, decoration = {calligraphic brace}]  (8.2,6.65) -- (8.2,8.65);
       \draw[-{Stealth}] (6.5,5.8)   -- (8,6.35);
      \draw [decorate, decoration = {calligraphic brace}]  (8.25,5.2) -- (8.25,7.5);
       \draw[-{Stealth}] (6.5,2.4)   -- (8,2.95);
      \draw [decorate, decoration = {calligraphic brace}]  (8.25,1.9) -- (8.25,4);
      \node[black] at (7.75,4.5) { $\vdots$};
      \node[black] at (8.38,4.8) {\scriptsize $\vdots$};
      \node[black] at (8.75,1) { $\textbf{z}_{t+1}$};
    \end{scope}
    \end{tikzpicture}
    \caption{\textit{A scheme of ADMM proposed in Sec. \ref{sec:proposedADMM}. Global vector $\textbf{z}_t$ is split into local copies $\tilde{\textbf{z}}^{(k)}_t$ used to constrain the estimate to $\textbf{x}^{(k)}_{t+1}$ (solving (\ref{eq:admm})). These estimates are again merged into a global variable $\textbf{z}_{t+1}$ and the procedure is repeated. Red is used to indicate the controllers shared among clusters, and yellow for the others.}}
    \label{fig:ADMMscheme}
\end{figure}

We now formulate a solution that includes coupling between the clusters, instead of solving each subproblem independently. In this way, we can produce a better estimate of the shared weights. For this, we apply the ADMM. In the workflow of ADMM, the objective function should be transformed to have the form
\begin{equation}
    \begin{split}
        \text{minimize   } & \Phi(\textbf{x}) + \Psi(\textbf{z}) \\
        \text{s.t.   }     & \textbf{Gx} + \textbf{Fz} = \textbf{c},
    \end{split}
\end{equation}
by choosing the functions $\Phi(\cdot)$ and $\Psi(\cdot)$ and the constraints. Similar to \cite{neumann2013sparse}, we dualize on the regularization term, i.e., we set $\Psi(\textbf{z}) = \alpha\textbf{1}^T\textbf{z}$, $\Phi(\textbf{x})=\|  f(\textbf{x})-\widehat{\textbf{b}} \|_2^2$, $\textbf{G} = \textbf{I}$, $\textbf{F} = -\textbf{I}$, and $\textbf{c} = \textbf{0}$. This corresponds to the general form consensus with regularization \cite{boyd2011ADMM}, with the following ADMM updates at each iteration $t+1$:

\begin{equation}\label{eq:admm}
    \begin{split}
        \textbf{x}_{t+1}^{(k)} & \in  \argmin_{\textbf{0}\leq\textbf{x}\leq\textbf{1}}\left(\|f^{(k)}(\textbf{x})-\widehat{\textbf{b}}^{(k)} \|_2^2 + \rho\| \textbf{x} - \tilde{\textbf{z}}^{(k)}_t + \textbf{u}^{(k)}_t\|_2^2 \right) \\
        \textbf{z}_{t+1} & = \textbf{S}^{-1}\left(\sum_{k=1}^K\textbf{q}^{(k)} - \frac{\alpha_k}{\rho}\right) \\
        \textbf{u}^{(k)}_{t+1} & =  \textbf{u}^{(k)}_t + \textbf{x}^{(k)}_{t+1} - \textbf{z}^{(k)}_{t+1}.
    \end{split}   
\end{equation}
A vector $\tilde{\textbf{z}}^{(k)}\in\mathbb{R}^{m_k}$ is a local copy of the global variable $\textbf{z}\in\mathbb{R}^m$, i.e., we have $(\tilde{\textbf{z}}^{(k)})_i = \textbf{z}_{\mathcal{G}(k,i)}$. The entries of $\textbf{q}^{(k)}\in\mathbb{R}^m$ are the values of $\textbf{x}^{(k)}_{t+1} + \textbf{u}^{(k)}_t$ obtained for the corresponding cluster, i.e., $(\textbf{q}^{(k)})_{\mathcal{G}(k,i)} = (\textbf{x}^{(k)}_{t+1})_i + (\textbf{u}_t^{(k)})_i$. Further, we proceed by solving the $x$-update step via coordinate descent, following the approach of \cite{rackovic2022CD}. The idea of ADMM is illustrated in Fig. \ref{fig:ADMMscheme}.


\section{Results}\label{sec:results}

\begin{figure*}
    \centering
    \includegraphics[width=\linewidth]{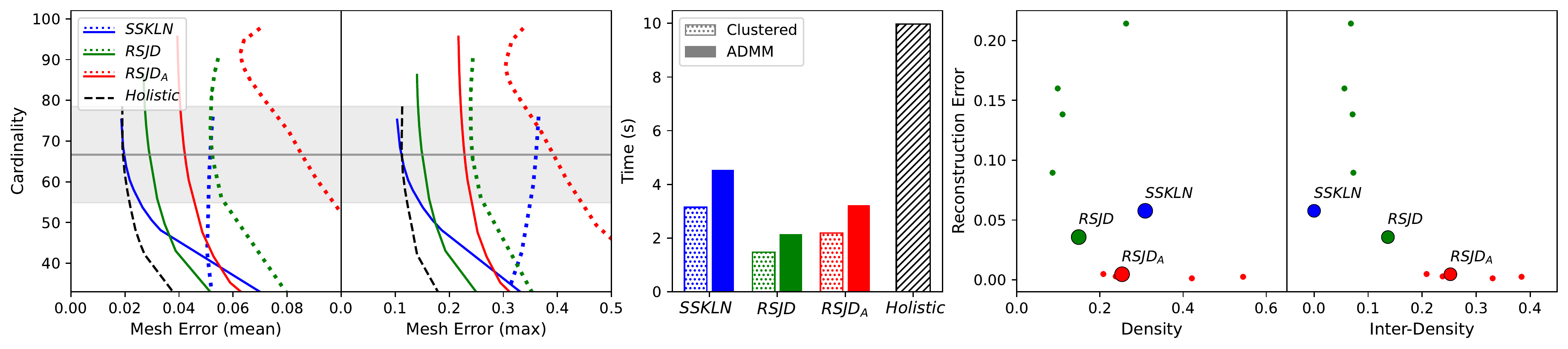}
    \caption{\textit{The first two subfigures show trade-off curves between the cardinality and mean and max RMSE, respectively, for different methods, as functions of the regularization parameter $\alpha>0$. The dotted lines represent a naive clustered solution, while the solid lines of the same color are the corresponding ADMM solution. The black dashed line shows a holistic approach. The gray horizontal line shows the cardinality of the ground-truth data, with a shaded region marking its standard deviation. The middle subfigure shows the average execution time per frame. The last two subfigures show trade-off curves between $E_R$ and $E_D$ and between $E_R$ and $E_{ID}$, respectively. Large dots indicate a chosen clustering, while the smaller ones of the same color are representing the discarded cases.}}
    \label{fig:fig_res}
\end{figure*}

For each of the four introduced clustering strategies (\textit{SSKLN, RSJD, RSJD\textsubscript{A}} and \textit{Sparse}), we will experiment with two possible approaches: 
\begin{enumerate}
    \item a naive clustered solution, where the subproblems (\ref{eq:clustered_objective}) are solved independently for each cluster, and in the end, the weights that are shared among multiple clusters are averaged;
    \item the proposed ADMM approach (\ref{eq:admm}), where the clusters can communicate the values of the shared weights, i.e., these components are constrained to be similar by coupling between the local and global variables, to get more confident estimates.
\end{enumerate}  Additionally, we include a holistic case, i.e., the method of \cite{rackovic2022CD}, where problem (\ref{eq:holistic_objective}) is solved without segmentation. 

Two main metrics of interest are mesh error and cardinality. Mesh error is computed as a root mean squared error (RMSE) between the target mesh $\widehat{\textbf{b}}$ and the estimated mesh $f(\hat{\textbf{w}})$, where $\hat{\textbf{w}}$ is the estimated weight vector $\text{RMSE}(\hat{\textbf{w}},\widehat{\textbf{b}}) =\sqrt{\frac{\|f(\hat{\textbf{w}})-\widehat{\textbf{b}}\|_2^2}{n}}.$ Cardinality is the number of non-zero weights in $\hat{\textbf{w}}$. An ideal solution should have low values for both of these.

The realistic, real-size human head character, \textit{Jesse}, used in our experiments, is publicly available within the MetaHumans platform  (\textcopyright unrealengine.com/en-US/eula/mhc). The avatar is manually animated by a human expert, to give a wide and realistic range of motion. Further, a small amount of Gaussian noise ($\sigma^2=0.03$cm as compared to the head width of $18$cm) is added to the mesh vertices, to mimic the realistic 3D scans used in the production. The model consists of $m=102$ blendshapes in the basis and $n=10000$ face vertices.

Initially, we need to choose a good value of the regularization parameter $\alpha$, for each approach, as well as the optimal $K$ for the \textit{RSJD} and \textit{RSJD}\textsubscript{A}. For this purpose, we run experiments on $300$ training frames with various values $\alpha>0$ and $0\leq K \leq m = 102$. The clusters for each of the approaches and choices of $K$ are shown in Fig. \ref{fig:MeshCLusters}. 

The results of the training are presented in Fig. \ref{fig:fig_TrainRes}. The left side of the figure gives trade-off curves between the mesh error and cardinality as a function of $\alpha$. Trade-off curves for a naive clustered solution are shown as dotted lines, while the same-color solid curves represent corresponding ADMM solutions. For the sake of visualization, the results of the four approaches are presented in separate subfigures. The gray horizontal line represents the average cardinality of the ground truth data, and the shaded region shows one standard deviation. We will mostly focus on the shaded area as it is indicated as a reasonable value for cardinality. Further, we will choose an optimal value of regularization $\alpha$ at which each curve crosses the gray line, as in this sense we will have a fair comparison of different methods.

Notice also that the results of the \textit{Sparse} approach are extremely poor, in most cases worse than a baseline which always predicts a zero weight vector (red vertical line). Hence, we will dismiss this approach from further consideration. Notice that in all the other cases, the results obtained using ADMM (solid curves) are outperforming those obtained via a naive clustered solution (dotted curves). For the \textit{RSJD} and \textit{RSJD}\textsubscript{A}, we should additionally pick an optimal choice of $K$. For this, we primarily look at the trade-off curves, but should also consider the execution time presented in the middle column of Fig \ref{fig:fig_TrainRes}. For both methods, $K=4$ leads to the best ADMM trade-off curve, however, the execution time is considerably longer for $K=4$ than for other choices. For the \textit{RSJD} we will choose $K=22$, as it gives only a slightly worse curve in the case of ADMM, while the execution time is almost half of the case with four clusters, and the results of a simple clustered method are actually the best performing for this choice. For \textit{RSJD}\textsubscript{A}, the relationship between the curves and the choice of $K$ follows the same pattern for both ADMM and a simple clustered approach, with an increase in $K$ leading to a decrease in the overall trade-off. We also notice that with $K=20$ the execution time is as low as it gets, hence we chose it as an optimal $K$. 

We take the selected set of results and present them together in Fig. \ref{fig:fig_res}. Notice that in all three methods, using ADMM significantly improves the results compared to the naive clustered approach. The trade-off curve of \textit{SSKLN} (using ADMM) is the only one reaching the accuracy of a holistic model, yet its execution time is the largest of the three distributed methods. An important observation is on the trade-off of $E_R$ versus $E_D$ and $E_{ID}$ (the last two subfigures). Here, the selected clusterings for each of the three methods, are presented with the annotated dots, while, for the \textit{RSJD} and \textit{RSJD}\textsubscript{A}, we additionally show the other five clusterings, as the same-color smaller dots. This confirms our assumptions from Sec. \ref{sec:choosingK}, that the cross-validation would lead to choosing \textit{RSJD} clustering with smaller $E_R$, and \textit{RSJD}\textsubscript{A} clustering with smaller $E_D$. Also, we can see a direct relationship between the execution time and $E_D$, as the clusterings with higher density lead to a longer execution.

\begin{figure}
    \centering
    \includegraphics[width=0.6\linewidth]{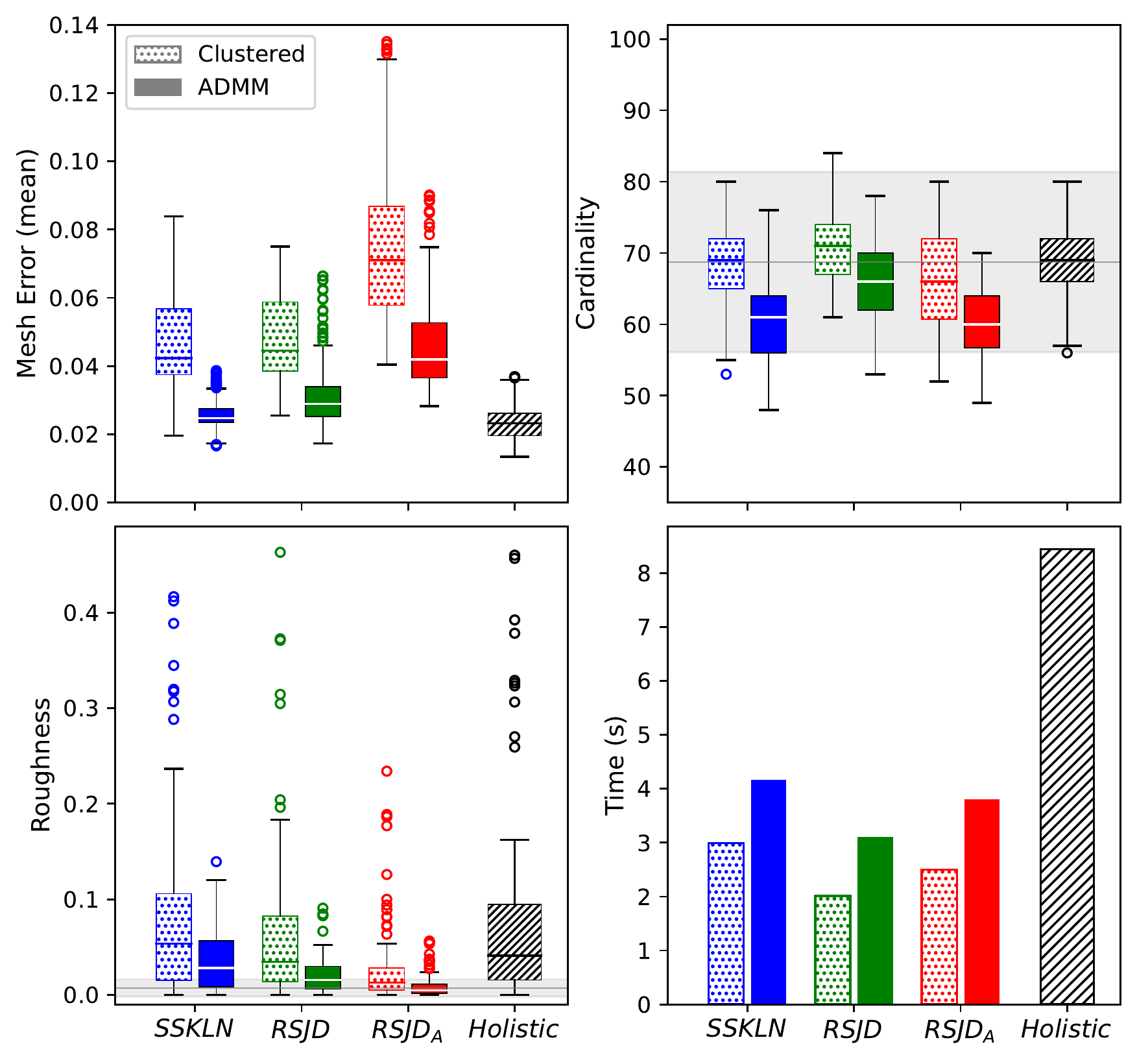}
    \caption{\textit{Results of the four methods over the test set. Dotted-face bars (and boxes) represent naive clustered solutions, while the same-color solid bars (and boxes) are the corresponding ADMM solutions. A gray horizontal line shows the metric value of the ground-truth data, and the shaded area gives one standard deviation.}}
    \label{fig:TestResults}
\end{figure}

Now we can observe the results of the test set in more detail. RMSE is presented in Fig. \ref{fig:TestResults} (upper left), where the solid-color boxes correspond to ADMM and dotted ones to a naive clustered solution. Like in the training set, a clear distinction between the two is also visible here --- in all three cases, the upper quartile of the ADMM solution is lower than the lower quartile of clustered solution. ADMM under the \textit{SSKLN} is comparable to the holistic in terms of median and quartiles, while ADMM under the \textit{RSJD} is just slightly worse. On the other side, the execution time of the clustered solution is lower than that of ADMM, as expected due to lack of cluster coupling, yet the difference is not as large as between the holistic case to others (Fig. \ref{fig:fig_res}, bottom left). As expected, cardinalities within the test set are similar across all the cases, showing only a slight advantage of ADMM (Fig. \ref{fig:TestResults}, bottom right).  Finally, since the test set consists of an animation sequence (see supplementary video), we are also interested in the temporal smoothness of the produced animation. This can be computed using the second-order differences to get the roughness penalty $\text{Roughness}(\hat{w}_i) = \sum_{t=2}^{T-1} \left(\hat{w}_i^{(t-1)} - 2\hat{w}_i^{(t)} + \hat{w}_i^{(t+1)}\right)^2,$ for a blendshape weight $\hat{w}_i$ over $T$ animated frames. Lower values of \textit{Roughness} correspond to smoother curves. The metric values are shown in Fig. \ref{fig:TestResults} (lower left). The values for ADMM are significantly lower than the corresponding values for a naive clustered approach and also compared to the holistic case. This can be noticed in the supplementary video as well, as the produced animation (especially for the \textit{RSJD} and \textit{RSJD}\textsubscript{A} clusters) is very smooth.

Cardinality stays in the shaded region of a ground-truth standard deviation, for all the approaches, as expected. Yet, it is slightly lower for the ADMM approaches than the other.

We might conclude that the application of ADMM on the clustered face leads to significant outperformance compared to the previous approaches that solved each cluster independently and averaged the shared components. ADMM produces visibly lower values of each considered metric, with the exception of the execution time. However, the execution time under ADMM is still less than half of the holistic approach. While all of the three selected methods seem to perform well in our use case, one could argue that the \textit{SSKLN} gives slightly preferable results. However, it is also important to recall that the \textit{SSKLN} demands the mesh clusters to be defined manually, hence it might be a less favorable choice. The \textit{RSJD} is slightly less accurate than the \textit{SSKLN}, but has better smoothness and the lowest execution time, as the clustered matrix is very sparse. 


\section{Conclusion}\label{sec:conclusions}

In this paper, we proposed a method for solving the inverse rig problem in a distributed manner. It is performed over the segmented animated face, applying the ADMM in order to include coupling between the clusters. Previously, the approaches with clustering would assume segments to be independent while fitting and averaging the shared components afterward. ADMM allows us to estimate the shared blendshapes jointly and hence get better estimates of the corresponding weights. Our method is general in the sense that it can work with different clustering approaches, as illustrated in this paper. We point out that, as the method is model-based, it makes sense to be applied with the data-free clustering methods, like the \textit{RS} \cite{romeo2020data} or \textit{RSJD} \cite{rackovic2021Clustering}, or the one proposed here; although other choices are also feasible. Irrespective of the clustering strategy used, applying the ADMM leads to improvements in all the metrics compared to a naive clustering scheme, except in the execution time. The differences are also visible in the supplementary video material, strongly favoring the ADMM solution.

We also propose an adjustment to the model-based clustering method of \textit{RSJD}, which applies a different strategy in assigning the blendshapes to mesh segments. The proposed method leads to increased density compared to the \textit{RSJD}, yet small in comparison to a holistic case, and often smaller than that of the \textit{SSKLN} \cite{seol2011artist}. Added complexity leads to an increased execution time (still almost half of the holistic approach) but to smoother and sparser results. 

Finally, we also propose a heuristic for choosing a good number of clusters $K$ in a data-free fashion. It is based on the trade-off between the density of the clustered blendshape matrix, and the reconstruction error. While these two are, in general, inversely proportional, producing a large number of clusterings with different $K$'s or initializations will point out the tendency of the results.\\

\textbf{Video Materials}

\url{https://youtu.be/fQaFA8CH2S4}\\

\textbf{Funding}

This work has received funding from the European Union's Horizon 2020 research and innovation program under the Marie Skłodowska-Curie grant agreement No. 812912, from FCT IP strategic project NOVA LINCS (FCT UIDB/04516/2020) and project DSAIPA/AI/0087/2018. The work has also been supported in part by the Ministry of Education, Science and Technological Development of the Republic of Serbia (Grant No. 451-03-9/2021-14/200125).

\bibliographystyle{unsrtnat}
\bibliography{references}  






\end{document}